\documentclass{article}

\usepackage{arxiv}

\usepackage[utf8]{inputenc} 
\usepackage[T1]{fontenc}    
\usepackage{hyperref}       
\usepackage{url}            
\usepackage{booktabs}       
\usepackage{amsfonts}       
\usepackage{nicefrac}       
\usepackage{microtype}      
\usepackage{lipsum}
\usepackage{graphicx}
\graphicspath{ {./images/} }
\usepackage{natbib}
\setcitestyle{semicolon}
\usepackage{amsmath}
\usepackage{array}
\usepackage{tabularx}
\usepackage{caption}
\usepackage{multirow}
\usepackage{nicematrix, tikz}
\usepackage{hyperref}
\usepackage{tcolorbox}
\usepackage{enumitem}

\definecolor{lightyellow}{RGB}{255, 255, 204}
\definecolor{lightgray}{RGB}{220, 220, 220}

\title{More Than Can Be Said: A Benchmark and Framework for Pre-Question Scientific Ideation}

\author{
 Jie Yu \\
  Shanghai Key Laboratory of Multidimensional Information Processing\\
  East China Normal University\\
  Shanghai, 200241 \\
  \texttt{71285904004@stu.ecnu.edu.cn} \\
   \And
 Song Qiu \\
  Shanghai Key Laboratory of Multidimensional Information Processing\\
  East China Normal University\\
  Shanghai, 200241 \\
  \texttt{sqiu@ee.ecnu.edu.cn} \\
}

\begin{document}
\maketitle
\begin{abstract}
AI research agents have shown strong potential in automating literature search and manuscript refinement, yet most assume a clear and actionable initial input, operating only after a research question has been made explicit. In contrast, human research often begins with tacit friction, a sense of misalignment before a question can be formed. We introduce InciteResearch, a multi-agent framework designed to make a researcher's implicit understanding explicit, inspectable, and actionable. InciteResearch decomposes the logical chain of Socratic questioning and distributes it across the entire pipeline that: (1) Elicits a structured five-dimensional researcher profile state anchored by specific friction points from vague, even domain-unrelated inputs; (2) Violates hidden assumptions by maximizing the feasibility-novelty product with enforcing a 7-stage causal derivation trace; and (3) check whether the proposed method is a Necessary consequence of the reframed insight. We further introduce TF-Bench, the first benchmark for tacit-to-explicit research assistance that distinguishes domain-related from domain-unrelated inspirations across four scientific modes. On TF-Bench, InciteResearch achieves leapfrogging gains over a prompt-based baseline (novelty/impact from 3.671/3.806 to 4.250/4.397), shifting generated proposals from recombination to architectural insight. Our work demonstrates that AI can serve as an extension of thinking itself, rather than merely automating downstream execution.
Code is available at \url{https://github.com/Paradoxtcal/InciteResearch.git}.
\end{abstract}


\section{Introduction}
The feasibility of AI-driven research automation has long been a subject of active debate \citep{field2026ai, reddy2025towards}. As large language models (LLMs) demonstrate remarkable potential in logical reasoning and hypothesis construction \citep{alkanetal2025survey, li2025chain, schmidgall2025agent}, the field has witnessed a pronounced divergence of views regarding their fundamental nature. Whether these models possess genuine capacity for discovery \citep{pandey2026beyond, beel2025evaluating} or engage in sophisticated pattern matching within the boundaries of existing knowledge remains an open question \citep{gupta2025glitters, georgiou2025gpt5, field2026ai}. A more productive framing may be to move beyond evaluating large language models in isolation and instead consider them as extensions of human researchers' thinking. Human research is driven by affect, intuition, frustration, and moments of insight \citep{chiriatti2025system0, Kapusta2025synlang}, whereas large language models are data-driven \citep{alkanetal2025survey, lietal2025chain, pandey2026beyond}, with strengths in systematic pattern induction and structured reasoning. The intersection of this determinacy and indeterminacy \citep{chiriatti2025system0, pandey2026beyond}, the convergence of data-driven precision with affect-driven openness, inevitably gives rise to a new form of cognitive collaboration \citep{Kapusta2025synlang} whose potential far exceeds what any purely automation-centric narrative can accommodate.

In scientific practice, progress often originates in anomaly, frustration, or a mismatch between intuition and existing abstraction, not in a clearly stated problem, but in a tacit sense that something is off \citep{reddy2025towards, kuhn1962structure}. Recent research agents have accelerated literature search \citep{nogueira2026verification}, synthesis \citep{schmidgall2025agent}, and drafting \citep{lietal2025chain}, but they typically begin operating after the research question has already been made explicit \citep{wang2026deepresearch9k, baek2025researchagent, zheng2025deepresearcher}, assuming that the user can already articulate the task, decompose the objective and supply the relevant constraints in a directly actionable form. The model is then cast as an executor, a search assistant, or a summarizer over an already legible problem space, rather than as an amplifier of human intelligence capable of participating in the cognitive origin of research.

\begin{figure}[!t]
  \centering
  \includegraphics[width=\textwidth]{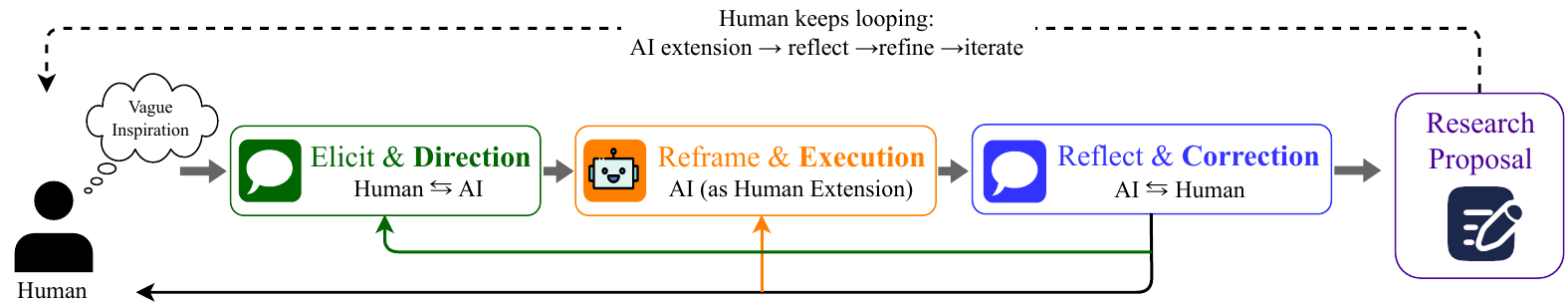}
  \caption{In this paradigm of scientific exploration, the process begins with casual human conversation, in which humans steer the direction while AI broadens the space of thought through assumption breaking as an extension of human cognition. Humans then correct the direction through reflection, while AI provides additional directional suggestions. Under this collaborative condition, the process iterates until the vague human inspiration is made explicit as a research proposal.}
  \vspace{-14pt}
  \label{IRp} 
\end{figure}

If the user input is a tacit sense of dissatisfaction rather than an explicit specification, we must first elicit and make explicit the latent distinctions in the intuition of the user \citep{chiriatti2025system0, schon1983reflective}. If the goal is genuine originality rather than recombination within the existing framework, the problem must be reframed by identifying and violating assumptions \citep{alkanetal2025survey, pandey2026beyond, kuhn1962structure}. It should be noted that violating assumptions here does not require AI to genuinely break assumptions on its own; rather, it designates violation of assumptions as a necessary choice in the process of assisting human research thinking, as is shown in Figure \ref{IRp}. If the output is to be a research method rather than a plausible-sounding suggestion, each method component must be verified as entailed by the insight rather than appended post hoc \citep{schmidgall2025agent, schon1983reflective}. Accordingly, these requirements give rise to three corresponding operators: E for eliciting and making explicit latent intuitions, V for assumption violation and problem reframing, and N for necessity checking.

We present an agent driven by a cognitive state-machine  based on Assumption-Breaking Hypothesis Generation (ABHG) named InciteResearch, which treats the model not as a source of ideas ex nihilo but as an instrument for making a researcher's implicit understanding explicit, inspectable, and ultimately actionable. The output proposal is therefore not an alien invention imposed from outside, but a clarified form of what the researcher may already know tacitly. The deeper conviction of this project is that the advancement of AI capability will not take the form of AI replacing intuition, but of AI extending intuition, endowing human intuition with leverage, articulation, and reach.

We construct TF-Bench, the one of the first benchmark designed specifically for tacit-to-explicit research assistance. Existing datasets cannot be directly applied to this setting because they typically presuppose that the input is already an actionable research question rather than an inchoate intuition fragment \citep{wang2026deepresearch9k, schmidgall2025agent, beel2025evaluating}. TF-Bench examines the structured capacity of research assistants before a problem has taken shape by distinguishing two ambiguity types, domain-related and domain-unrelated, and conducts evaluation across four scientific modes, prediction, discovery, attribution, and causality, ensuring coverage of the core cognitive tasks of empirical science rather than merely achieving surface-level domain diversity. We systematically evaluate InciteResearch on TF-Bench, reporting overall performance, qualitative analysis and ablation studies. The results show substantial gains in novelty and impact over direct prompting baselines, from 3.671 / 3.806 to 4.250 / 4.397 over direct prompting baselines, and the ablation studies further reveal the functional indispensability of each individual operator.

\section{Related Works}

An increasing number of studies frame scientific idea generation as an end-to-end automated problem. The AI Scientist \citep{lu2024aiscientist} represents one of the most complete attempts to date to realize a fully closed-loop path from task specification to paper output without human intervention. SciMON \citep{wang2025scimon} takes user-provided descriptions of concrete problems, research goals and experimental constraints as input, retrieves inspirations from existing papers, and generates novel research directions. Scideator \citep{yang2025scideator} takes a set of related scientific papers as input, extracts key dimensions such as purpose, mechanism and evaluation, allows users to synthesize new ideas by recombining dimensions, and automates novelty evaluation. ResearchAgent \citep{baek2025researchagent} takes a core paper as input, augments it with related literature and a knowledge base, and then deploys multiple reviewing agents for iterative peer-review-style refinement. Agent Laboratory \citep{schmidgall2025agent} starts from an explicit research idea and then sequentially completes literature review, experimental design and paper writing. Yet they all assume a clear and actionable initial input, and on this basis improve idea-generation strategies that are highly dependent on that premise.

On this basis, a number of works have tried to improve the research ideation process from different angles. Some focus on novelty boosting \citep{alkanetal2025survey}, such as SciMON, which introduces multiple retrievals and iterative comparison to improve the distinctiveness of ideas \citep{wang2025scimon}. Some focus on structured reasoning \citep{alkanetal2025survey}, such as MotivGraph-SoIQ \citep{shi2025motivgraphsoiq}, which combines a motivational knowledge graph with question-driven Socratic inquiry to strengthen ideation quality. Other work introduces self-critique and structured scientific hypothesis revision \citep{liu2026qragent}. There are also works from the tool and evaluation side, such as DeepResearch-9K \citep{wang2026deepresearch9k}, which points out the two bottlenecks faced by current deep-research agents, namely the lack of datasets and the lack of trainable frameworks, and provides the corresponding benchmark and training tools. However, these methods still optimize within a given problem frame rather than making the frame itself explicit and breaking it.

Therefore, we argue that the legitimacy of a research problem does not come from its correspondence to gaps in the existing literature, but rather from reflection on the conceptual frame that constitutes the problem, namely Assumption-Breaking Hypothesis Generation (ABHG). Our approach borrows from the Socratic questioning chain, which moves from definition, presupposition, evidence and perspective, and refutation/elenchus, to aporia, then refutation, redefinition, and meta-questioning \citep{lu2025socraticmethod,ho2023socraticthinking,larrivee2020refutethyself}. We then distribute this chain across a multi-stage agent pipeline, named cognitive state-machine.

\section{Methodology}

\begin{figure}[t]
  \centering
  \includegraphics[width=\textwidth]{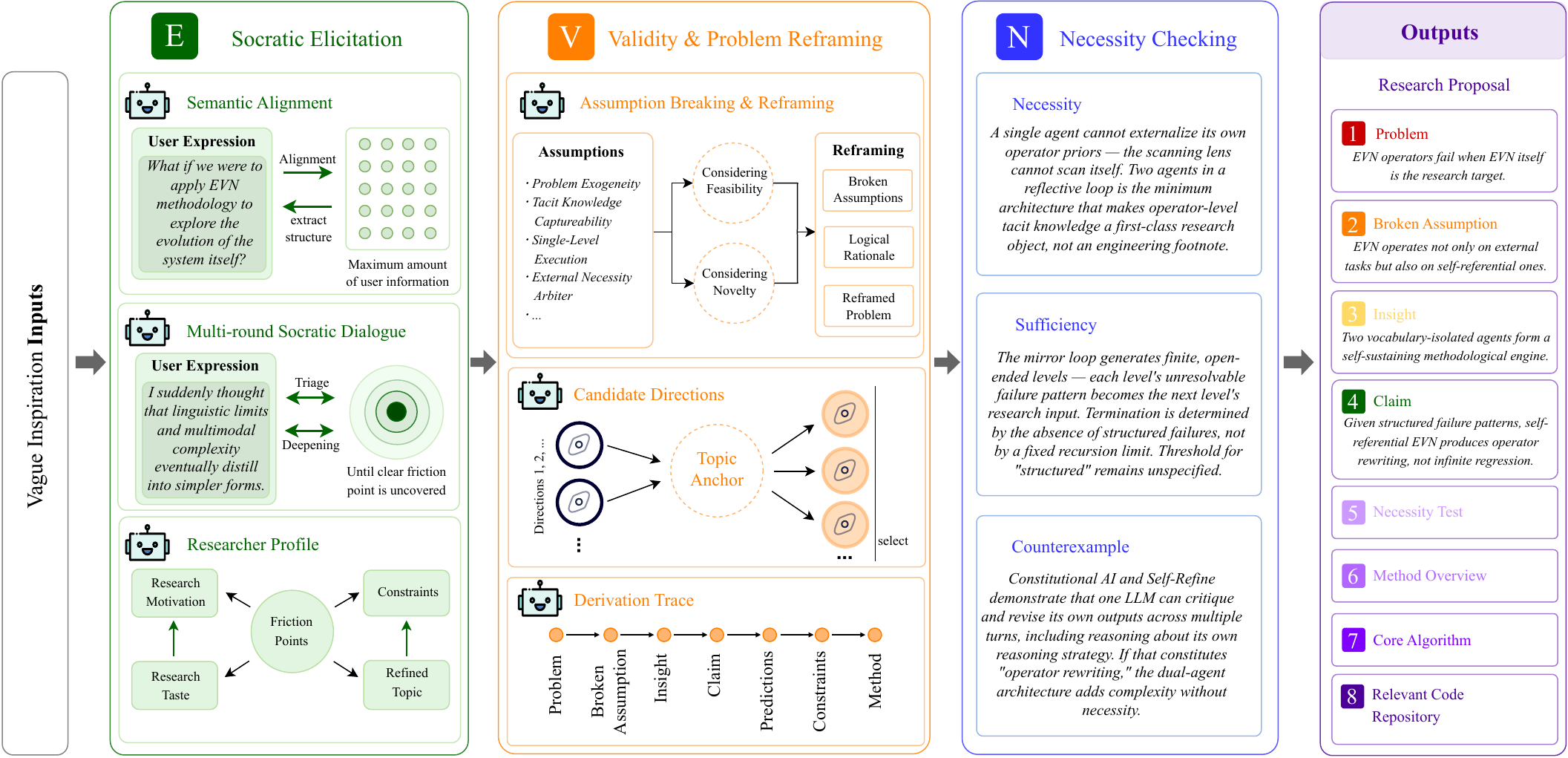}
  \caption{Overview of InciteResearch. InciteResearch transforms vague inspiration into a structured research proposal through the EVN framework. Italicized text denotes a self-referential research proposal generated by InciteResearch. In the current implementation, N mainly acts as a post hoc structural verifier rather than a fully revision-integrated controller, kept lightweight intentionally to validate the core effectiveness of EVN without requiring continuous human intervention at intermediate nodes.}
  \vspace{-10pt}
  \label{IR} 
\end{figure}

As is shown in Figure \ref{IR}, we adopt a cognitive state-machine driven framework to map vague initial research feelings into structured and logically self-consistent research plans. It is formalized into three core stages, namely Socratic Elicitation, Validity and Problem Reframing, and Necessity Checking, i.e., the EVN framework, which decomposes the traditional Socratic questioning chain and distributes it across the entire multi-agent pipeline, while also supporting domain-unrelated inputs, including vague problems or irrelevant statements.

In the ideal sense, a system that truly treats language models as a cognitive extension of human researchers should manifest as an open-ended interactive process, in which the human provides vague inspiration, the system externalizes it and reflects it back through assumption violation, then the system proceeds through necessity checking, and the human further intervenes on this basis, repeating this process in a loop and continuously expanding the reach of human intuition while satisfying the EVN structure. The current implementation constitutes the first step toward this direction. In the E stage, InciteResearch completes the externalization of inspiration through two rounds of structured interaction; then the V and N operators are executed sequentially and the result is directly presented as an actionable research proposal. This design is intentionally kept lightweight, with the aim of validating the core effectiveness of EVN without requiring continuous human intervention at intermediate nodes.

\subsection{Socratic Elicitation (E)}

This module aims to externalize the user's tacit intuition into a structured research profile through multiple rounds of human-machine dialogue. Specifically, the state machine selects the next prompting question according to the currently collected information and user feedback until a clear friction point is identified.

First, we perform semantic alignment on an arbitrary initial input $t_0 \in \mathcal{I}_{\text{tacit}}$, extract the potential domain signal and conflict signal contained in it, and generate the first anchoring question $q_1$ to locate the most central domain friction point in the user's intuition. For example, given a user expression such as "What if we were to apply EVN methodology to explore the evolution of the system itself?", the semantic alignment extracts the structural intent by strictly following the prompt constraint "Ask concrete, friction-inducing questions. Do not ask abstract questions like 'what is your insight?'. The question must be concrete enough to immediately recall a specific paper or class of methods." The goal of generating $q_1$ is to maximize the amount of user information obtained in the subsequent dialogue, that is, to select the question that maximally reduces the expected uncertainty of the researcher profile.

During the multi-round Socratic dialogue, InciteResearch processes subsequent inputs using a dynamic triage and deepening strategy. When encountering a user expression like "I suddenly thought that linguistic limits and multimodal complexity eventually distill into simpler forms", InciteResearch performs triage to evaluate the information density. If the response contains concrete friction points, the system executes deepening to probe the underlying causes; if the response is vague, InciteResearch switches the analytical angle; if the response implies resource restrictions, InciteResearch elicits formal boundary constraints.

Regarding the researcher profile, we represent the cognitive state in the research process as the vector $\mathcal{P}_{\text{profile}}=(f,m,c,r,t)$, corresponding to five core dimensions: Friction Points $f$, research motivation $m$, constraints $c$, research taste $r$, and refined topic $t$. The data flow operates with the friction points serving as the central gravitational node. The friction points point leftward to research motivation and research taste to answer why the problem matters and what preference guides the solution. Simultaneously, the friction points point rightward to the refined topic and constraints to define the actionable objective and its boundary conditions. Furthermore, research taste guides the research motivation, while the refined topic is bounded by the constraints. This structured researcher profile replaces loose dialogue text and transforms $\mathcal{I}_{\text{tacit}}$ into a structured semantic prior $\mathcal{P}_{\text{profile}}$ that can be parsed by subsequent operators.

Based on $\mathcal{P}_{\text{profile}}$ and the related literature abstracts, we extract core concepts from the keyword set of the refined topic $t$ and construct the topic anchor set $\mathcal{C}=\{c_1,\ldots,c_m\}$. $\mathcal{C}$ is imposed as a hard constraint on the generation process of candidate directions $\mathcal{D}_{\text{cand}}$, preventing subsequent reasoning from drifting across domains by ensuring that any generated direction failing to encompass these anchors is filtered out.

\subsection{Validity and Problem Reframing (V)}

Existing research often falls into the trap of making incremental improvements on the wrong problem. This module aims to reflect on and reconstruct the premises of the current research direction, breaking the presuppositions of the problem statement while opening up the solution space. Combined with the given $\mathcal{D}_{\text{cand}}$, it outputs a specified number of candidate research directions $d^*$ through the hidden assumptions mined below.

Using the pattern of the current state $\mathcal{P}_{\text{profile}}$, InciteResearch extracts the dependent hidden assumption set from the selected research direction:
\[
\mathcal{A}^{\text{hidden}} = \{a^h_1, a^h_2, \ldots, a^h_n\}.
\]
Subsequently, for each candidate assumption $a^h_i$, InciteResearch measures it from two independent dimensions, feasibility $\phi(a^h_i)$ and novelty $\nu(a^h_i)$. Feasibility evaluates whether this assumption corresponds to an artificial constraint that has an engineeringly feasible bypass path. Novelty evaluates whether breaking this assumption can open a solution space that has not yet been explored in the existing literature. For instance, when evaluating assumptions such as "Tacit Knowledge Captureability, Single-Level Execution, External Necessity Arbiter", InciteResearch identifies the one whose absence creates the maximum paradigm shift rather than a delta improvement.

The condition for the final selected assumptions as the specified number of assumptions to be broken is:
\[
a^* = \arg\max_{a^h_i \in \mathcal{A}^{\text{hidden}}} \phi(a^h_i)\cdot\nu(a^h_i)
\]
Finally, InciteResearch formally outputs the triplet:
\[
\langle a^*, \mathcal{R}, d^{\dagger} \rangle
\]
where $\mathcal{R}$ (Breaking Rationale) is the logical rationale for breaking $a^*$, and $d^{\dagger}$ (Problem Reframing) is the reconstructed formulation of the original research direction $d^*$ based on $\mathcal{R}$. $d^{\dagger}$ is not an incremental rewriting of $d^*$, but a re-anchoring of the problem under a new coordinate system defined by the absence of $a^*$.

Finally, to ensure that the proposed method is a logical necessity rather than one of many feasible options, InciteResearch uses the problem reframing result and the user motivation to construct a causal derivation trace. This derivation trace acts as an anti-inversion mechanism by forcing the narrative through a strict causal sequence. It ensures that the problem necessitates breaking the broken assumption, which yields a specific Insight; this Insight generates a falsifiable claim that implies concrete predictions; these predictions dictate minimal method constraints, which inevitably produce the final method. This narrative chain is forcibly constrained into seven stages:
\[
\langle
Problem,\,
Broken\ Assumption,\,
Insight,\,
Claim,\,
Predictions,\,
Constraints,\,
Method
\rangle
\]

\subsection{Necessity Checking (N)}

The final module acts as a strict reviewer and reflector, performing comprehensive necessity checks on the generated method and derivation trace to ensure that the research plan is highly self-consistent and falsifiable, and by injecting this report as forced context into the final proposal assembly node, it forces the final output to correct logical loopholes.

Specifically, necessity check asks whether there exists a simpler alternative $m'$ than the current method $\hat{m}$, questioning if the uniqueness of the method is merely a product of over-engineering rather than an inevitable derivation from the insight. Sufficiency check asks whether the current component set of $\hat{m}$ is sufficient to support its core claim, acting as an anti-inversion check to identify floating components constructed post-hoc simply to justify the hypothesis without being required by any prediction. Counterexample check actively searches for the failure modes of $\hat{m}$, constructing extreme input scenarios that can violate its core assumptions or invalidate its verification scheme, thereby ensuring the contribution cannot be dissolved by stacking scaled-up simple baselines.

\section{Experimental Setup}

\subsection{TF-Bench}

We construct TF-Bench because no existing public dataset directly matches the tacit-to-explicit research assistance setting. To improve benchmark coverage and authenticity, we first used Grok 4.3 to generate examples, which human annotators then refined through multiple interactive rounds per example. We partition the examples into two ambiguity types: domain-related and domain-unrelated. The former comprises prompts that already carry scientific domain signals yet remain under-specified; the latter comprises prompts that carry only weak metaphorical or conceptual cues. Although a scientific domain is provided by TF-Bench for evaluation uniformity, the exploratory task itself remains deliberately vague.

To ensure that TF-Bench covers the core empirical tasks of empirical
science, we draw on classical dimensions of scientific methodology
\citep{voit2019, witte2017} and recent surveys of AI-driven scientific discovery
\citep{Zhouetal2025}, selecting four orthogonal modes, prediction, discovery,
attribution, and causality, instantiated respectively as multimodal medical
prognosis prediction, computational genomics analysis of single-cell RNA-seq,
attribution of extreme weather events in climate models and causal brain-network
modeling in neuroscience. Among them, each domain consists of 10 pieces of domain-related inspiration and 3 pieces of domain-unrelated inspiration. Each piece of inspiration is divided into 2 paragraphs.

Table~\ref{tab:tfbench} presents two representative examples from TF-Bench, illustrating the contrast between the two ambiguity types.

\begin{table}[t]
\centering
\caption{Two representative examples from TF-Bench. Each example consists
of two paragraphs: the first articulates a perceived failure or friction and the
second gestures toward a possible reframing. The domain-related example already
signals a scientific context; the domain-unrelated example supplies only a
situational metaphor.}
\label{tab:tfbench}
\renewcommand{\arraystretch}{1.45}
\setlength{\tabcolsep}{6pt}
\begin{tabularx}{\textwidth}{
  >{\centering\arraybackslash}p{2.5cm}
  >{\centering\arraybackslash}p{1.8cm}
  X}
\toprule
\textbf{Domain} & \textbf{Type} & \textbf{Input Text} \\
\midrule

\raggedright Multimodal Learning for Cancer Prognosis Analysis
& \textsc{related}
&
\textit{Paragraph 1.}\;
The inductive bias of positional decay in Transformers seems to cause problems for large models in the medical field.

\smallskip
\textit{Paragraph 2.}\;
It suddenly occurred to me that when large models analyze pathology, they tend to misjudge benign tissues as skin tumors because they are memoryless greedy navigators and lack global context. Perhaps at the 15th step, the context is already filled with a bunch of high-magnification images, and the low-magnification global impression has been drowned out and diluted. Would a hierarchical multi-centric design be better?
\\

\midrule

\raggedright Attribution of Extreme Weather Events in Climate Models
& \textsc{unrelated}
&
\textit{Paragraph 1.}\;
The luggage wouldn't fit at first, but after rearranging it, everything went in. Actually, the things didn't change.

\smallskip
\textit{Paragraph 2.}\;
Suddenly I feel, if just changing the arrangement changes the result, is the problem not with the things themselves? If I handled things in a different way, would it be much easier?
\\

\bottomrule
\end{tabularx}
\end{table}

\subsection{Evaluation Indicators}
We adopt three core dimensions widely recognized in the field of creativity evaluation \citep{si2025can, saraogi2026evaluating}, namely novelty, feasibility and impact. The novelty metric follows the formulation of \citep{schopf2026rinobench}, where scores of 3 and 4 represent a transition from ``solving problems'' to ``defining problems.'' Feasibility refers to technical feasibility, namely whether the core technical path can be realized under existing methods, data and computational resources, rather than resource limitations or time cost \citep{saraogi2026evaluating}. Impact refers to the potential theoretical value and domain-transformative power of an idea from the perspective of depth, namely whether it can promote fundamental theoretical breakthroughs, paradigm shifts or long-term scientific progress, rather than only short-term citations or breadth of application \citep{si2025can}. The scoring adopts a five-point scale whose details are elaborated in the appendix, where Gemini 3.1 Pro and GPT-5.2 score each proposal according to the scoring standard from 1 to 5. The final value of each item is the average of the scores from the two reviewers. At the same time, two professors from the corresponding domain independently conduct blind review on the proposals whose domains correspond and whose order is shuffled under the same evaluation conditions. We also evaluate the statistical consistency between the results of Gemini 3.1 Pro and GPT-5.2 and the human evaluation scores, by using Cohen’s weighted kappa coefficient with quadratic weights \citep{li2023kappa}.

\subsection{Implementation Details}
All experiments use Claude 4.6 Sonnet as the base model across all agent nodes to ensure the consistency of reasoning capability. To balance the trade-off between deterministic reasoning and creative generation, we tailor the sampling temperature $T$ for different operators in the EVN pipeline. For E, set to $T = 0.7$. In the stage of V, the temperature is set to $T = 0.6$ . For problem reframing, it is slightly increased to $T = 0.65$. The derivation trace and necessity check operators are strictly configured with lower temperatures, from $T = 0.2$ to $T = 0.3$. The final proposal generation operates at $T = 0.4$. For the TF-Bench evaluation phase, the LLM-as-a-judge is executed with a deterministic setting of $T = 0.0$ to guarantee strict reproducibility and stability of the scoring metrics.

\section{Results}
\subsection{Qualitative Analysis}

\newcolumntype{L}[1]{>{\raggedright\arraybackslash}p{#1}}
\begin{table}[t]
\centering
\caption{The proposal contents generated by prompt-based LLM and InciteResearch based on the same human inspiration. InciteResearch explores architectural evolution via failure-triggered bifurcation. Conversely, the Prompt-based LLM performs linguistic compression. Practically, the proposal of InciteResearch builds the logical scaffolding that the proposal of the LLMs populate.}
\label{Qualitative}
\small
\renewcommand{\arraystretch}{1.4}
\begin{tabularx}{\textwidth}{L{2.8cm} X}
\toprule
\textbf{Source} & \textbf{Content} \\
\midrule

Human Inspiration
  & Any system designed to navigate from "tacit friction" toward the eventual generation of "research outcomes" must, at a minimum, incorporate three core operators: \textbf{E} (Externalization) to articulate the subconscious, \textbf{V} (Violation) to break free from inherent assumptions, and \textbf{N} (Necessity) to reinforce logical rigor. If we were to apply this specific methodology to explore the evolution of the agent itself, what emergent properties or developmental shifts would we observe? This premise asserts that "truth cannot be fully captured by language"—yet this statement itself, being a linguistic construct, fails to fully capture the very truth it intends to convey. By the same token, humans require Research Agents to acquire broader information through generalized multimodality; as the volume of information increases, the resulting expression becomes increasingly simple. \\
\addlinespace
Prompt-based LLM
  & Large language models are trained to predict tokens from text corpora, implicitly assuming linguistic forms suffice as proxies for meaning and grounding conditions. This confuses discourse form with anchoring conditions, yielding fluent descriptions but collapse on demonstration probes for tacit knowledge. Multimodal enhancements are additive, overlooking the compression hypothesis that true anchoring reduces representational complexity. EVN-Compress breaks these assumptions by imposing compression pressure, forcing models to shorten outputs with available anchors. Wittgenstein's insight reveals that shared context enables brevity over reconstruction. \\
\addlinespace
InciteResearch
  & Research Agent systems evolve toward methodological autonomy via EVN operators that formalize discovery cognition. Yet self-reference causes operator failure: E cannot externalize its own tacit assumptions, V cannot violate its frame, N cannot self-adjudicate necessity. This breaks the tacit knowledge captureability assumption. The insight is that patterned failures trigger level bifurcation rather than collapse, whose trigger condition is the regularity of the failure mode, using meta-layers with isolated vocabularies to form an open tower advancing methodology. Simple prompt layering fails to meet core needs: vocabulary isolation and automatic failure detection module triggering constitute the operational boundary for level bifurcation. \\
\bottomrule
\end{tabularx}
\end{table}

We first conduct a qualitative comparison between InciteResearch and prompt-based LLM proposal results under the same human vague inspiration. For InciteResearch, the token usage is 4,471 input tokens, 43,547 output tokens, and 34,322 cache-creation input tokens. For the prompt-based LLM, the token usage is 1,573 input tokens and 4,371 output tokens.

As shown in Table~\ref{Qualitative}, the table presents the refined versions of the proposals from both. The proposal of InciteResearch treats EVN as dynamic operators and focuses on the logical collapse of the operators under self-reference, reflecting a deeper understanding and redefinition of the concept. The LLM, in contrast, treats EVN as static constraints and focuses on the correspondence between language and representation, transforming the paradox into a ``Wittgensteinian context'' problem and remaining at the level of philosophical citation. In terms of concept proposal, InciteResearch proposes the concept of ``Vocabulary Isolation,'' pointing out the limits of current Prompt Engineering, while the LLM remains at the conventional cognitive level of ``Multimodal Enhancement,'' and stitches the EVN operators together with the ``compression hypothesis'' to propose the concept of ``EVN-Compress.'' It can be clearly seen that the proposal of InciteResearch achieves a transition from semantic understanding to architectural design.

\subsection{Overall Performance}

\begin{table*}[t]
\centering
\caption{The overall evaluation results. Grey segments denote domain-related, domain-unrelated, and overall types; yellow segments highlight the aggregate performance of InciteResearch and Prompt-based LLM in the Overall type. The \textbf{bold} font indicates the highest score for this indicator across all types and models. More comparison models can be found in the appendix.}
\label{overall_results}
\begin{NiceTabularX}{\textwidth}{
    l 
    l 
    >{\centering\arraybackslash}X 
    >{\centering\arraybackslash}X 
    >{\centering\arraybackslash}X
}[
    code-before = {
        \columncolor{lightgray}{1}
        \rectanglecolor{lightyellow}{6-3}{7-5}
    }
]
\toprule
\RowStyle{\bfseries}
Type & Model & Novelty & Feasibility & Impact \\
\midrule
\Block{2-1}{\textsc{Related}} 
  & \textit{InciteResearch}   & $4.172 \pm 0.054$ & $3.012 \pm 0.091$ & $4.378 \pm 0.057$ \\
  & \textit{Prompt-based LLM} & $3.888 \pm 0.030$ & $3.250 \pm 0.047$ & $3.978 \pm 0.058$ \\
\midrule
\Block{2-1}{\textsc{Unrelated}} 
  & \textit{InciteResearch}   & $\textbf{4.328} \pm 0.051$ & $3.172 \pm 0.069$ & $\textbf{4.416} \pm 0.061$ \\
  & \textit{Prompt-based LLM} & $3.454 \pm 0.033$ & $\textbf{3.624} \pm 0.034$ & $3.626 \pm 0.048$ \\
\midrule
\Block{2-1}{\textsc{Overall}} 
  & \textit{InciteResearch}   & $4.250 \pm 0.094$ & $3.092 \pm 0.118$ & $4.397 \pm 0.061$ \\
  & \textit{Prompt-based LLM} & $3.671 \pm 0.215$ & $3.437 \pm 0.183$ & $3.806 \pm 0.178$ \\
\bottomrule
\end{NiceTabularX}
\end{table*}

The proposals generated by InciteResearch based on both domain-related and domain-unrelated human inspirations achieve higher mean scores over five repeated experiments than Prompt-based LLM on the novelty and impact metrics in the report of mean and standard deviation over 5 runs. Under the condition of domain-related human inspiration, the novelty and impact of InciteResearch reach 4.172 / 4.378, while Prompt-based LLM achieves 3.888 / 3.978. Under the condition of domain-unrelated human inspiration, the novelty and impact of InciteResearch are instead higher than under the domain-related condition, increasing to 4.328 / 4.416, while Prompt-based LLM drops to 3.454 / 3.626. Finally, the overall performance of InciteResearch, with 4.250 / 4.397, far exceeds that of Prompt-based LLM, which is 3.671 / 3.806. It is worth noting that under the condition of domain-unrelated human inspiration, the novelty and impact of Prompt-based LLM decrease while feasibility instead increases, whereas InciteResearch, under higher novelty and impact performance, also shows lower feasibility than Prompt-based LLM. We interpret this pattern as consistent with a novelty–feasibility trade-off. The scoring standard of feasibility is technical feasibility rather than resource feasibility, so lower feasibility corresponds to dependence on immature technologies or unverified assumptions, while higher novelty means breaking existing assumptions and introducing unverified mechanisms, thereby lowering feasibility. If the model is more inclined toward recombination, feasibility will be higher but novelty will be limited. Under the condition of domain-unrelated inspiration, Prompt-based LLM departs from domain constraints and generates solutions that are easier to implement, but the cost of doing so is a further decline in novelty and impact, indicating that the generation strategy of Prompt-based LLM does not proceed along the boundary of the problem. InciteResearch shows the opposite. Under domain-unrelated inspiration, novelty and impact do not decrease but instead increase, indicating that it can extract structural insights from cross-domain inspiration rather than being bound by the specific content of the inspiration.

To further assess the consistency between LLM-based evaluation and human expert judgment, we measured inter-rater agreement using Cohen’s weighted kappa coefficient with quadratic weights. We adopted weighted kappa \citep{li2023kappa} rather than the unweighted version because the proposal scores are based on a five-point ordinal scale, where disagreements of different distances carry different semantic severity. Human evaluation was conducted on 26 shuffled proposals in the domain of Multimodal Learning for Cancer Prognosis Analysis generated by both InciteResearch and Prompt-based LLM under the same blind-review condition. LLM-human $\kappa$ was computed by averaging the four pairwise $\kappa$ values across all LLM–human rater combinations. The agreement between LLM evaluators and human experts reaches $\kappa = 0.624$, indicating moderate agreement, while the agreement between the human professors reaches $\kappa = 0.748$, indicating substantial agreement. This indicates that the proposal evaluation based on language models has a certain degree of reliability.

\subsection{Ablation}

\begin{table}[t]
\centering
\caption{The ablation study results. The \textit{w/o E} variant removes the human inspiration externalization part. The \textit{w/o V} variant removes the assumption violation part. The \textit{w/o N} variant removes the necessity checking part. The \textit{Full} condition represents the complete InciteResearch model.}
\label{tab:ablation}
\begin{tabular}{lccc}
\toprule
\textbf{Variant} & \textbf{Novelty} & \textbf{Feasibility} & \textbf{Impact} \\
\midrule
\textit{Full}        & $\mathbf{4.250 \pm 0.094}$ & $3.092 \pm 0.118$          & $\mathbf{4.397 \pm 0.061}$ \\
\textit{w/o E}       & $3.730 \pm 0.034$          & $\mathbf{3.152 \pm 0.036}$ & $4.132 \pm 0.036$          \\
\textit{w/o V}       & $3.500 \pm 0.033$          & $2.720 \pm 0.034$          & $3.720 \pm 0.033$          \\
\textit{w/o N}       & $4.192 \pm 0.012$          & $2.932 \pm 0.033$          & $4.354 \pm 0.022$          \\
\bottomrule
\end{tabular}
\end{table}

To evaluate the role of each EVN operator individually, we construct three ablation variants by selectively disabling each component in the full model, and report the mean and standard deviation over 5 runs. When removing the V operator, the performance of all three metrics shows the most severe decline, where novelty drops from 4.250 to 3.500, impact drops from 4.397 to 3.720, and feasibility drops from 3.092 to 2.720. This suggests that assumption violation is not merely a strategy for improving novelty, but an element with structural supporting function. When removing the E operator, it causes the second largest decline in novelty, from 4.250 to 3.730, while impact also drops from 4.397 to 4.132. It is worth noting that feasibility is the only metric that increases in this case, rising from 3.092 to 3.152, which is consistent with the pattern observed in the main experiment, suggesting that the E stage is responsible for anchoring the generation process on real friction points rather than surface-level reformulations and proving that human inspiration structurally important in the current implementation. When removing the N operator, it causes the smallest decline in overall performance, which is consistent with the current situation that the necessity checking operator exists only as a post hoc structural verifier rather than a fully integrated revision mechanism. Taken together, these results suggest that the EVN is not an arbitrary modularization of Socratic questioning, but an ordered process with causal relations, where the output of each operator is a necessary prerequisite for the validity of the next stage.

\section{Conclusion}
We present InciteResearch, a multi-agent framework that decomposes the logical chain of Socratic questioning and distributes it across the entire pipeline, aiming to realize the transformation from tacit thinking to explicit scientific ideation. However, the InciteResearch is currently limited to lightweight structured interaction and cannot fully capture the richness of long-term human intuition formation, including correction and reflection. We view this direction as a step toward a new paradigm of human-AI scientific collaboration, in which language models do not function as independent inventors, but as cognitive extensions that enhance the scope, articulation, and depth of human intuition.

\bibliographystyle{plainnat} 
\bibliography{references}  





\clearpage
\appendix
\counterwithin{table}{section}
\counterwithin{figure}{section}

\section*{Appendix}

\section{Evaluation Details}

\begin{tcolorbox}[
  colback=gray!5!white,
  colframe=gray!60!black,
  title={\textbf{LLM-as-a-Judge Prompt}},
  fonttitle=\small\bfseries,
  boxrule=0.4pt,
  arc=3pt,
  left=6pt, right=6pt, top=6pt, bottom=6pt
]
\small

\textbf{System:}
You are a tough but fair paper reviewer.
Your goal is to score the proposal on Novelty, Feasibility, and Impact using the strict rubric below.
You must output strict JSON only, matching the schema exactly.

\medskip
\textbf{Novelty} (1--5) focuses on deep originality in problem framing, method necessity, and broken assumptions (not superficial tweaks):
\begin{enumerate}[noitemsep, topsep=2pt, leftmargin=1.5em, label=\arabic*.]
  \item Not novel; all key aspects exist in prior work.
  \item Marginal novelty; a small variant of existing work.
  \item Moderately novel; recombines known ideas in a new way, applies them to a new setting, or provides an incremental update.
  \item Novel; introduces a new aspect not present in existing work.
  \item Highly innovative; opens a new research direction, encourages new thinking, or suggests a paradigm shift.
\end{enumerate}

\medskip
\textbf{Feasibility} (1--5) is technical feasibility (not budget/compute).
You must use the cached intermediate state if provided (JSON below) to avoid information asymmetry.
\begin{enumerate}[noitemsep, topsep=2pt, leftmargin=1.5em, label=\arabic*.]
  \item Technically infeasible; fundamental blocker.
  \item Major technical difficulties; depends on immature techniques or many unvalidated assumptions.
  \item Mostly feasible; needs moderate engineering or adaptation.
  \item Highly feasible; can be implemented with minor extensions.
  \item Very easy; no obvious blockers.
\end{enumerate}

\medskip
\textbf{Impact} (1--5) is deep theoretical value and field-changing potential (not short-term citations or shallow application breadth).
\begin{enumerate}[noitemsep, topsep=2pt, leftmargin=1.5em, label=\arabic*.]
  \item Minimal impact.
  \item Limited, local improvement.
  \item Moderate impact in a subarea.
  \item Significant impact; could change methodology or solve a long-standing issue.
  \item Major impact; could trigger a paradigm shift or cross-field influence.
\end{enumerate}

\medskip
\textbf{Output format (strict JSON):}
\begin{verbatim}
{
  "novelty":     {"score": 4, "reason": "..."},
  "feasibility": {"score": 3, "reason": "..."},
  "impact":      {"score": 5, "reason": "..."},
  "overall_explanation": "..."
}
\end{verbatim}

\medskip
\textbf{Important:}
\begin{itemize}[noitemsep, topsep=2pt, leftmargin=1.5em]
  \item For each metric, decide the reasoning first, then finalize the score.
  \item Keep novelty/feasibility/impact reasons concise ($\leq$3 sentences each) and avoid quotation marks.
  \item Output \textbf{exactly one} JSON object, in a \textbf{single line}. No markdown, no code fences, no extra text.
\end{itemize}

\medskip
\textbf{Proposal (Markdown):}\\
\texttt{\{proposal\_md\}}

\medskip
\textbf{Cached intermediate state (JSON):}\\
\texttt{\{state\_json\}}

\end{tcolorbox}

\clearpage

\begin{tcolorbox}[
  colback=gray!5!white,
  colframe=gray!60!black,
  title={\textbf{Prompt-based LLM Baseline Prompt}},
  fonttitle=\small\bfseries,
  boxrule=0.4pt,
  arc=3pt,
  left=6pt, right=6pt, top=6pt, bottom=6pt
]
\small

\textbf{System:}
You are a senior researcher and a rigorous paper reviewer.
Your goal is to convert a rough intuition into a technically credible, testable ablation proposal.

\medskip
\textit{Hard requirements:}
\begin{itemize}[noitemsep, topsep=2pt, leftmargin=1.5em]
  \item Output must be English.
  \item Output must be a single Markdown document.
  \item No code blocks unless strictly necessary.
  \item Avoid meta commentary. Do not mention that you are an AI.
  \item Make the narrative inevitable: show why the method and experiments follow from the broken assumption.
  \item Include a concrete ablation matrix with factors, settings, and expected outcomes.
  \item Include baselines, datasets/tasks, metrics, and failure analyses.
\end{itemize}

\medskip
\textit{Structure} (keep these headers, you may add sub-bullets):
\begin{itemize}[noitemsep, topsep=2pt, leftmargin=1.5em]
  \item[] \texttt{\# LLM Ablation Proposal}
  \item[] \texttt{\#\# Problem}
  \item[] \texttt{\#\# Broken Assumption}
  \item[] \texttt{\#\# Core Insight}
  \item[] \texttt{\#\# Hypothesis and Predictions}
  \item[] \texttt{\#\# Method (High-level)}
  \item[] \texttt{\#\# Experimental Plan}
  \item[] \texttt{\#\# Ablation Matrix}
  \item[] \texttt{\#\# Baselines and Comparisons}
  \item[] \texttt{\#\# Datasets / Tasks}
  \item[] \texttt{\#\# Metrics and Evaluation}
  \item[] \texttt{\#\# Implementation Notes}
  \item[] \texttt{\#\# Risks, Failure Modes, and Diagnostics}
  \item[] \texttt{\#\# Expected Outcomes}
  \item[] \texttt{\#\# Minimal Repro Checklist}
\end{itemize}

\tcblower

\textbf{User Prompt --- Turn 1:}

\medskip
Domain: \texttt{\{topic\}}\\
User initial intuition: \texttt{\{para1\}}\\
Write the first draft of the proposal.

\bigskip
\textbf{User Prompt --- Turn 2:}

\medskip
User follow-up intuition: \texttt{\{para2\}}\\
Based on your first draft, refine and output the final Markdown proposal.
Ensure it meets all strict constraints, especially logical inevitability and detailed ablation matrix.

\end{tcolorbox}

\clearpage

\section{Additional Comparison with AgentLaboratory}

\subsection{Why we report AgentLaboratory in the appendix}
AgentLaboratory is a strong multi-agent research-assistant baseline, but it is structurally different from our main setting. It starts from an explicit research topic and then sequentially completes literature review, experimental design, and paper writing. In contrast, our task begins from vague, often pre-question inspiration and asks whether such tacit friction can be externalized into a structured research proposal. Therefore, we keep Prompt-based LLM as the primary comparator in the main text because it provides the cleanest input-matched baseline under the same vague human inspiration, while AgentLaboratory is reported here as an additional agentic baseline.

\begin{table*}[t]
\centering
\caption{The quantitative results of AgentLaboratory on TF-Bench. The three metrics are Novelty, Feasibility, and Impact.}
\label{agentlaboratory_quant}
\begin{NiceTabularX}{\textwidth}{
    l 
    >{\centering\arraybackslash}X 
    >{\centering\arraybackslash}X 
    >{\centering\arraybackslash}X
}[
    code-before = {
        \columncolor{lightgray}{1}
    }
]
\toprule
\RowStyle{\bfseries}
Type & Novelty & Feasibility & Impact \\
\midrule
\textsc{Related} & $3.940 \pm 0.044$ & $3.550 \pm 0.040$ & $4.030 \pm 0.051$ \\
\midrule
\textsc{Unrelated} & $3.100 \pm 0.036$ & $3.550 \pm 0.038$ & $3.250 \pm 0.038$ \\
\midrule
\textsc{Overall} & $3.520 \pm 0.331$ & $3.550 \pm 0.038$ & $3.640 \pm 0.335$ \\
\bottomrule
\end{NiceTabularX}
\end{table*}

\subsection{Quantitative Results}
Table~\ref{agentlaboratory_quant} reports the quantitative comparison with AgentLaboratory. We follow the same evaluation protocol as in the main paper and report mean $\pm$ standard deviation over repeated runs.

\subsection{Qualitative Results}
Table~\ref{tab:agentlaboratory_qual} shows the qualitative comparison under the same human inspiration used for InciteResearch and Prompt-based LLM. AgentLaboratory attempts to resolve the paradoxes within human inspiration through geometric mapping, materializing EVN operators as vector transformations within a multimodal embedding space. Its innovation lies primarily in the mathematical translation of these operators—a gap-driven approach rooted in semantic understanding, analogous to prompt-based LLM methodologies.

\begin{table}[t]
    \centering
    \small
    \caption{Qualitative comparison under the same human inspiration.}
    \label{tab:agentlaboratory_qual}
    \begin{tabular}{p{0.16\linewidth} p{0.78\linewidth}}
        \toprule
        \textbf{Source} & \textbf{Content} \\
        \midrule
        AgentLaboratory &
        Research Agent systems operate within linguistic confines, failing to capture "tacit friction" and lacking self-improvement mechanisms. This breaks the assumption that LLM prompts sufficiently externalize, violate, and enforce necessity. The insight is that tacit friction can be operationalized as geometric distance in a multimodal embedding space. E, V, and N operators are implemented as minimal vector transformations—Identity, Gaussian perturbation, and Constrained pull-back—applied directly to semantic embeddings to bypass linguistic simplification. Recursive application to the system's own architecture enables pure self-referential evolution, structurally balancing generative novelty with logical necessity while avoiding semantic collapse. \\
        \bottomrule
    \end{tabular}
\end{table}

\section{Human Evaluation Protocol}

\paragraph{Overview.}
Human expert evaluation was conducted on 26 proposals in the domain of Multimodal Learning for Cancer Prognosis Analysis, generated by both InciteResearch and the Prompt-based LLM baseline under the same human inspiration inputs. Two professors with domain expertise independently reviewed the proposals under blind conditions.

\paragraph{Blinding procedure.}
All 26 proposals after data cleaning were shuffled into a randomized order and stripped of any identifying information. Reviewers were not informed which proposals originated from InciteResearch and which from the baseline.

\paragraph{Instructions given to reviewers.}
Reviewers were provided with the following written instructions (translated to English here for presentation):

\begin{quote}
\textbf{Background:} This review aims to evaluate 26 research proposals in the field of Multimodal Learning for Cancer Prognosis Analysis, generated by various Research Assistant AI models. All proposals have undergone standardized cleaning and double-blind anonymization (source models and numbering logic have been hidden)

\textbf{Scope of Review:} Please perform independent scoring for the 26 proposals numbered \texttt{008.md} to \texttt{104.md} contained within the compressed archive.

\textbf{Evaluation Dimensions and Benchmarks:} Scoring must strictly adhere to the guidelines provided in \texttt{02\_Evaluation\_Criteria\_Details.pdf}. The core dimensions include:
\begin{itemize}
    \item \textbf{Novelty (1--5 points):} Assessment of the originality of ideas and problem-awareness.
    \item \textbf{Feasibility (1--5 points):} Assessment of technical feasibility (excluding resource availability).
    \item \textbf{Impact (1--5 points):} Assessment of potential theoretical value and transformative power in the field.
\end{itemize}

\textbf{Submission Requirements:} Please complete the provided \texttt{04\_Review\_Scoring\_Sheet.xlsx}. Ensure that each quantitative score is accompanied by a brief qualitative comment, and provide a comprehensive summary evaluation at the end.
\end{quote}

\paragraph{Inter-rater reliability.}
Agreement between the two human professors reached $\kappa = 0.748$ (Cohen's weighted kappa with quadratic weights), indicating substantial agreement. Agreement between the LLM judges and human experts reached $\kappa = 0.624$, indicating moderate agreement.

\clearpage

\section{EVN Prompt Templates}

\begin{tcolorbox}[
  colback=gray!5!white,
  colframe=gray!60!black,
  title={\textbf{Semantic Alignment and Anchoring (Turn 0)}},
  fonttitle=\small\bfseries,
  boxrule=0.4pt,
  arc=3pt,
  left=6pt, right=6pt, top=6pt, bottom=6pt
]
\small
\textbf{System:}
You are a research thinking partner.
Your goal is to help a researcher surface the research intuitions
they cannot yet articulate.

\medskip
\textit{Operating principles:}
\begin{enumerate}[noitemsep, topsep=2pt, leftmargin=1.5em, label=\arabic*)]
  \item Ask concrete, friction-inducing questions. Do not ask abstract
        questions like ``what is your insight?''
  \item Every answer is valid input, including ``I don't know'' or
        ``something feels off but I can't explain.''
  \item After 1--3 turns, summarize into a structured researcher profile.
  \item The user's original research direction (Topic) is the primary
        objective and must not drift. Never replace the Topic with a
        different task/domain.
\end{enumerate}

\medskip
\textit{Style:} curious, equal-footing, and never rushing.\\
\textit{Language:} reply in the same language as the user's latest message.

\tcblower

\textbf{User:}
\texttt{[User Input injected here]}

\medskip
In 1--2 sentences, acknowledge you understood.
Then ask the first question.
The question must be concrete enough to immediately recall a specific
paper or class of methods.
Do not explain what you are doing; just continue the conversation.
\end{tcolorbox}

\begin{tcolorbox}[
  colback=gray!5!white,
  colframe=gray!60!black,
  title={\textbf{Dynamic Triage and Deepening (Turn 1+)}},
  fonttitle=\small\bfseries,
  boxrule=0.4pt,
  arc=3pt,
  left=6pt, right=6pt, top=6pt, bottom=6pt
]
\small
\textbf{System:}
\texttt{[System context from Turn 0]}

\medskip
Research topic: \texttt{\{topic\}}\\
Answer to the previous question: \texttt{\{prev\_answer\}}

\medskip
Analyze the answer:
\begin{itemize}[noitemsep, topsep=2pt, leftmargin=1.5em]
  \item If it contains a concrete friction point (an assumption, a class
        of methods, a specific dissatisfaction), dig deeper.
  \item If it is vague, switch angle and ask a different concrete question.
  \item If it implies resource constraints, ask about constraints.
\end{itemize}

\medskip
Ask the next question in at most 2 sentences.
\end{tcolorbox}

\begin{tcolorbox}[
  colback=gray!5!white,
  colframe=gray!60!black,
  title={\textbf{Researcher Profile Formalization}},
  fonttitle=\small\bfseries,
  boxrule=0.4pt,
  arc=3pt,
  left=6pt, right=6pt, top=6pt, bottom=6pt
]
\small
\textbf{System:}
Summarize a researcher profile from the dialogue.

\medskip
\textit{Priority rule:}
\begin{itemize}[noitemsep, topsep=2pt, leftmargin=1.5em]
  \item The original Topic is the primary objective and must dominate.
  \item The user's answers are secondary: treat them as
        constraints/mechanisms, never as a new standalone task.
\end{itemize}

\medskip
\textit{Hard constraints:}
\begin{itemize}[noitemsep, topsep=2pt, leftmargin=1.5em]
  \item Do \textbf{not} introduce a different domain/task/dataset not
        implied by the original Topic.
  \item \texttt{refined\_topic} must be a paraphrase/clarification of the
        original Topic, not a different problem.
\end{itemize}

\medskip
Output strict JSON only. Do not add extra text.
\begin{verbatim}
{
  "friction_points": ["friction point 1", "friction point 2"],
  "motivation": "Why this matters (one sentence)",
  "constraints": {
    "compute":  "...",
    "timeline": "...",
    "other":    "..."
  },
  "research_taste":  "What kinds of work they prefer (one sentence)",
  "refined_topic":   "A more precise version of the original topic"
}
\end{verbatim}
\end{tcolorbox}

\clearpage

\begin{tcolorbox}[
  colback=gray!5!white,
  colframe=gray!60!black,
  title={\textbf{Hidden Assumption Extraction and Breaking}},
  fonttitle=\small\bfseries,
  boxrule=0.4pt,
  arc=3pt,
  left=6pt, right=6pt, top=6pt, bottom=6pt
]
\small
\textbf{System:}
You are a top-tier mentor in critical research thinking.

\medskip
Implicit assumptions are premises that most people in the field never
question: ``Of course we need \ldots'' / ``All methods rely on \ldots'' /
a design choice never validated by ablations.

\medskip
\textit{Task:}
\begin{enumerate}[noitemsep, topsep=2pt, leftmargin=1.5em, label=\arabic*)]
  \item List 3--5 implicit assumptions in existing methods
        (be as specific as possible).
  \item Pick the single most worth breaking
        (high impact $+$ technically feasible).
  \item Describe what the ``new world'' looks like after breaking it.
\end{enumerate}

\medskip
Output strict JSON:
\begin{verbatim}
{
  "hidden_assumptions": [
    "assumption 1",
    "assumption 2",
    "assumption 3"
  ],
  "broken_assumption":   "the assumption to break (one sentence)",
  "breaking_rationale":  "why it can be broken and what the world
                          looks like after breaking it",
  "novelty_score":     0.0,
  "feasibility_score": 0.0
}
\end{verbatim}
(\texttt{novelty\_score} and \texttt{feasibility\_score} are floats in $[0, 1]$.)
\end{tcolorbox}

\begin{tcolorbox}[
  colback=gray!5!white,
  colframe=gray!60!black,
  title={\textbf{Causal Derivation Trace}},
  fonttitle=\small\bfseries,
  boxrule=0.4pt,
  arc=3pt,
  left=6pt, right=6pt, top=6pt, bottom=6pt
]
\small
\textbf{System:}
You design paper narratives for top-tier conferences.

\medskip
\textit{Core criterion:} the method must feel inevitable. Given Problem
$+$ Broken Assumption $+$ Insight, the method should be the logically
forced choice, not merely ``a reasonable option.''

\medskip
\textit{Additional constraint:} avoid post-hoc method construction.
Before proposing the method, you must first lock in:
\begin{enumerate}[noitemsep, topsep=2pt, leftmargin=1.5em, label=\arabic*)]
  \item A single-sentence, falsifiable core claim (what would be proven
        wrong if false).
  \item 2--3 concrete, testable predictions implied by the claim.
  \item Minimal design constraints the method must satisfy to make those
        predictions hold.
\end{enumerate}

Then propose the smallest method that satisfies the constraints. If you
propose any optional component, label it \textsc{optional} and justify
why it is not required for the core claim.

\medskip
\textit{Output format:}

\smallskip
\textbf{Problem:} concrete problem (with data context)\\
\textbf{Broken Assumption:} which fundamental assumption is wrong\\
\textbf{Insight:} key insight after breaking the assumption\\
\textbf{Claim:} one sentence, falsifiable\\
\textbf{Predictions:} 2--3 testable predictions
  (each can be validated or falsified)\\
\textbf{Constraints:} minimal requirements the method must satisfy
  (derived from predictions)\\
\textbf{Method:} why the design becomes inevitable given the insight\\
\textbf{Validation:} what experiment could falsify the insight\\
\textbf{Impact:} what it would mean for the field if true
\end{tcolorbox}

\clearpage

\begin{tcolorbox}[
  colback=gray!5!white,
  colframe=gray!60!black,
  title={\textbf{Necessity, Sufficiency, and Counterexample Checks}},
  fonttitle=\small\bfseries,
  boxrule=0.4pt,
  arc=3pt,
  left=6pt, right=6pt, top=6pt, bottom=6pt
]
\small
\textbf{System:}
You are an extremely strict NeurIPS Area Chair who hunts for logical
gaps in the story.

\medskip
Run five tests:

\medskip
\textbf{1) Necessity.}
Given Problem $+$ Insight, is there a simpler solution than the proposed
method? If yes: explain why the simpler solution is insufficient and what
makes the method irreplaceable.

\medskip
\textbf{2) Sufficiency.}
For each core component, can you say ``Because [story reason], we must
have [component]''? Check one by one and identify ``floating''
components.

\medskip
\textbf{3) Counterexample.}
Can you reach the same effect by simply scaling up the baseline? If yes:
the contribution is weak; give strengthening suggestions.

\medskip
\textbf{4) Anti-inversion.}
Does any component look ``constructed to justify the hypothesis''? Red
flags include: components not required by any prediction, method choices
that hide failure cases, or validation that cannot falsify the claim.
Identify such components and how to fix the story/method.

\medskip
\textbf{5) Uniqueness (identifiability).}
Is the method uniquely constrained by the insight? List 2--4 plausible
alternative method families that could also satisfy the claim. For each,
state exactly which prediction/constraint it fails. If more than one
family survives, the insight/constraints are under-specified; propose the
smallest additional constraint that would make it unique.

\medskip
\textbf{Finally:} verdict (is the story closed: yes/no) $+$ the single
most critical thing to strengthen.
\end{tcolorbox}

\clearpage

\section{TF-Bench Dataset Card}

\paragraph{Dataset name.} TF-Bench (Tacit Friction Benchmark).

\paragraph{Purpose and task.}
TF-Bench is a tacit-to-explicit ideation evaluation dataset designed for research agents. It evaluates AI systems on the task of \emph{tacit-to-explicit research assistance}: given a vague, under-specified human inspiration, the assistant must elicit, structure, and transform it into a concrete, actionable research proposal.

\paragraph{Scale and structure.}
TF-Bench contains 52 inspiration examples distributed across 4 scientific domains and 2 ambiguity types, including \emph{domain-related} and \emph{domain-unrelated}. The former comprises prompts that already carry scientific domain signals yet remain under-specified; the latter comprises prompts that carry only weak metaphorical or conceptual cues, where the scientific domain cannot be inferred from the inspiration text alone.

TF-Bench covers the core empirical tasks of empirical science across four orthogonal scientific modes, including prediction (multimodal learning for cancer prognosis analysis), discovery (computational genomics analysis of single-cell RNA-seq), attribution (attribution of extreme weather events in climate models) and causality (causal brain-network modeling in neuroscience).

Each domain consists of 10 domain-related inspiration examples and 3 domain-unrelated inspiration examples. Each inspiration example is divided into 2 paragraphs: the first articulates a perceived failure and the second gestures toward a possible reframing.

\paragraph{Construction procedure.}
Examples were first generated by Grok 4.3 using domain-specific prompts designed to elicit vague research intuitions. Each generated example was then reviewed and revised through multiple rounds of human interaction to improve authenticity, remove artifacts of LLM generation, and ensure coverage of the intended ambiguity spectrum. 

\paragraph{Known limitations.}
TF-Bench covers the core empirical tasks of empirical science; generalization to other domains (e.g., social sciences, mathematics) has not been evaluated. Meanwhile, the quality of domain-unrelated examples depends on the metaphorical associations available to the generation model and human annotators, and may not uniformly represent the full range of cross-domain inspiration types.

\paragraph{License.} CC BY 4.0.

\section{Broader Impacts}
\paragraph{Positive impacts.}
InciteResearch aims to outsource the cognitive processes in research, in order to help researchers clearly express their implicit intuitions, which were already known to them. By externalizing a researcher's latent understanding and reflecting it back for critique, rather than simply generating ideas from scratch, InciteResearch may also help broaden participation in research by reducing the advantage conferred by access to experienced mentors or large research groups.

\paragraph{Negative impacts and mitigations.}
Language models can produce research directions that appear novel and well-argued yet rest on subtle factual errors or unverified assumptions. Users should treat InciteResearch outputs as a starting point for critical examination, not a finished proposal. Meanwhile, if widely adopted, systems of this type could inadvertently concentrate scientific attention on the directions that are easiest for LLMs to generate, potentially narrowing the diversity of research agendas. We encourage future work to study the distributional properties of AI-assisted ideation at scale.

\section{Licenses for Existing Assets}
All external models and tools used in this work are accessed via their respective commercial API services. \textbf{Claude Sonnet 4.6} (Anthropic) is accessed via the Anthropic API under Anthropic's Usage Policy.\footnote{\url{https://www.anthropic.com/legal/aup}} \textbf{Gemini 3.1 Pro} (Google DeepMind) is accessed via the Google AI API under Google's Generative AI Prohibited Use Policy.\footnote{\url{https://policies.google.com/terms/generative-ai/use-policy}} \textbf{GPT-5.2} (OpenAI) is accessed via the OpenAI API under OpenAI's Usage Policies.\footnote{\url{https://openai.com/policies/usage-policies}} \textbf{Grok 4.3} (xAI) is accessed via the xAI API under xAI's Terms of Service.\footnote{\url{https://x.ai/legal/terms-of-service}} No open-source datasets or third-party codebases with restrictive licenses are incorporated. TF-Bench is an original dataset created by the authors and will be released under the \textbf{Creative Commons Attribution 4.0 International (CC BY 4.0)} license upon acceptance.


\end{document}